\documentclass[runningheads]{llncs}
\usepackage[T1]{fontenc}
\usepackage{graphicx}
\graphicspath{{figures/}}
\usepackage{booktabs}
\usepackage{amsmath}
\usepackage{amsfonts}
\usepackage{caption}
\usepackage{url}
\usepackage{siunitx}
\usepackage[utf8]{inputenc}
\usepackage{amsmath, amssymb}
\usepackage{enumitem}
\usepackage{listings}
\usepackage{DejaVuSans} 

\usepackage[misc]{ifsym}
\newcommand{\corr}{(\Letter)}
\usepackage{mwe}
\usepackage{xcolor}
\usepackage[numbers]{natbib} 

\usepackage{lscape}

\usepackage{lipsum}

\newcommand\blfootnote[1]{%
  \begingroup
  \renewcommand\thefootnote{}\footnote{#1}%
  \addtocounter{footnote}{-1}%
  \endgroup
}

\begin{document}

\title{DriftMoE:
A Mixture of Experts Approach to Handle Concept Drifts}

\titlerunning{DriftMoE: 
A Mixture of Experts Approach to Handle Concept Drifts} 

\authorrunning{Aspis, Cajas Ordoñez, et al.}

\author{Miguel Aspis*\inst{1}\orcidID{{0009-0005-7560-513X}} 
\and 
Sebasti\'an A. Cajas Ordoñez*\inst{1}\orcidID{0000-0003-0579-6178} \corr \and \\
Andr\'es L. Su\'arez-Cetrulo\inst{1} \orcidID{0000-0001-5266-5053} \and
Ricardo Sim\'on Carbajo\inst{1}\orcidID{0000-0002-2121-2841}}

\institute{Ireland's National Centre for Artificial Intelligence (CeADAR) \\ 
University College Dublin, Belfield, Dublin, D04 V2N9 
Ireland’s Centre for Applied AI (CeADAR) \\
University College Dublin, Belfield, Dublin, D04 V2N9, Ireland\\
\email{\{miguel.aspis1, sebastian.cajasordonez, andres.suarez-cetrulo, ricardo.simoncarbajo\}@ucd.ie}}

\maketitle              

\blfootnote{* The first two authors have equally contributed to the development of this work.}

\begin{abstract}
Learning from non-stationary data streams subject to concept drift requires models that can adapt on-the-fly while remaining resource-efficient. Existing adaptive ensemble methods often rely on coarse-grained adaptation mechanisms or simple voting schemes that fail to optimally leverage specialized knowledge. This paper introduces DriftMoE, an online Mixture-of-Experts (MoE) architecture that addresses these limitations through a novel co-training framework. DriftMoE features a compact neural router that is co-trained alongside a pool of incremental Hoeffding tree experts. The key innovation lies in a symbiotic learning loop that enables expert specialization: the router selects the most suitable expert for prediction, the relevant experts update incrementally with the true label, and the router refines its parameters using a multi-hot correctness mask that reinforces every accurate expert. This feedback loop provides the router with a clear training signal while accelerating expert specialization. We evaluate DriftMoE's performance across nine state-of-the-art data stream learning benchmarks spanning abrupt, gradual, and real-world drifts testing two distinct configurations: one where experts specialize on data regimes (multi-class variant), and another where they focus on single-class specialization (task-based variant). Our results demonstrate that DriftMoE achieves competitive results with state-of-the-art stream learning adaptive ensembles, offering a principled and efficient approach to concept drift adaptation. All code, data pipelines, and reproducibility scripts are available in our public GitHub repository:  \url{https://github.com/miguel-ceadar/drift-moe}.
\color{black}
\end{abstract}

\section{Introduction}

Real-world and production data architectures are characterized by the proliferation of high-velocity data streams, generated continuously by an ever-expanding network of sources, including IoT sensors, financial tickers, social media feeds, and network monitoring systems \cite{gama2014survey}. Unlike static datasets, these streams are inherently non-stationary; their underlying data-generating process can and often does change over time. This phenomenon, known as concept drift \cite{Tsymbal2004TheWork}, invalidates the core assumption of classical machine learning that training and test data are drawn from the same distribution. Consequently, predictive models trained on historical stream data can suffer model decay when the concept they have learned shifts. This challenge is stressed in applications spanning the edge-to-cloud continuum \cite{10.1145/3721889.3721929}, where resource-constrained edge devices must perform real-time inference and adapt to local environmental changes without constant retraining from a central cloud, making on-the-fly adaptation a critical requirement \cite{ordonez2024adaptive}.

Learning in dynamic, streaming environments requires a departure from traditional batch-learning paradigms. The field of online incremental learning addresses this by processing data one instance at a time, continuously updating the model as new information arrives. However, this introduces a challenge known as the stability-plasticity dilemma \cite{mermillod2013stability}. A model must be plastic enough to quickly adapt to new, evolving concepts but also stable enough to retain previously acquired knowledge that may still be relevant, preventing catastrophic forgetting. Achieving this balance is the central goal of online incremental learning algorithms designed for concept-drifting data streams.

To address this challenge, a predominant and successful strategy for tackling concept drift is the use of adaptive online ensembles \cite{Gomes2017AdaptiveClassification, Brzezinski2014OAUE}. Drawing inspiration from batch ensemble methods, these approaches maintain a collection of diverse base learners, typically incremental decision trees like the Hoeffding Tree \cite{domingos2000mining}. Their strength lies in their modularity and inherent ability to adapt. As surveyed in \cite{SUAREZCETRULO2023118934}, these ensembles typically employ one of two strategies: active or passive. Active methods explicitly use a change detection algorithm (e.g., ADWIN \cite{bifet2007learning}) to signal a drift, triggering a mechanism to adapt the ensemble, such as resetting the worst-performing learner. Passive methods, on the other hand, adapt continuously through mechanisms like dynamic weighting or a sliding window over the data. Models like Adaptive Random Forest (ARF) \cite{Gomes2017AdaptiveClassification} and Leveraging Bagging \cite{bifet2010leveraging} have become state-of-the-art benchmarks by demonstrating robust performance across a wide variety of drift scenarios.

Despite their success, current adaptive ensemble approaches face several limitations \cite{SUAREZCETRULO2023118934, gama2014survey}. Active methods rely heavily on drift detection algorithms, which can suffer from false positives or delayed detection, leading to suboptimal adaptation timing \cite{bifet2007learning}. When drift is detected, the typical response, such as resetting underperforming learners, is often coarse-grained and reactive, with performance decay until the change is detected \cite{gama2014survey}. Passive methods typically use simple mechanisms that may not optimally leverage the specialized knowledge of individual learners. Furthermore, most existing approaches lack a principled mechanism for experts to develop specialized knowledge for different data regimes or concepts, limiting their ability to handle complex, multi-faceted drift scenarios efficiently.

To address these limitations, we propose DriftMoE, a novel architecture that reframes the adaptation mechanism through the lens of a Mixture of Experts (MoE). Instead of relying on explicit drift detectors or simple voting schemes, DriftMoE features a lightweight neural network router that is co-trained alongside a pool of incremental decision tree experts. The router dynamically assigns incoming data instances to the most suitable experts, while both components adapt continuously through a symbiotic online training loop. This integrated approach offers a more nuanced method for managing expert contributions and specialized knowledge in non-stationary environments.

The principal novelty of our approach lies in its symbiotic, fully online training loop between the router and the experts. We explore two distinct configurations for expert specialization: i) a multi-class approach where experts learn different data regimes, and ii) a task-based approach where each expert specializes in a single class. For each incoming data instance, the router dynamically assigns the instance to the most suitable expert. Once the true label is revealed, the relevant experts (Top-K in the multi-class variant, all in the task variant) update themselves incrementally. The router then refines its own parameters in small online batches using a novel training signal: a multi-hot correctness mask that positively reinforces every expert that predicted the instance correctly, creating a cooperative feedback mechanism. As experts specialize and become more accurate on specific data regimes or tasks, they provide a clearer training signal to the router. In turn, as it gets smarter, the router improves at channeling the right data to the right expert, accelerating specialization and improving the overall predictive performance of the model. This tight integration contrasts with traditional ensembles, where adaptation is often a more coarse and reactive process.


This paper is organized as follows. Section 2 reviews related work on adaptive ensembles and MoE architectures. Section 3 details our approach and online training procedure. Section 4 presents the experimental setup and results. Section 5 discusses key findings, and Section 6 concludes with future directions.



\section{Related Work}

Our research lies at the intersection of two well-established fields: online incremental learning on data streams and MoE architectures. We first review the dominant ensemble-based approaches for handling concept drift and then discuss the MoE paradigm, highlighting the novelty of applying it in a fully online, streaming context.

\subsection{Adaptive Ensembles for Concept Drift}

The challenge of learning in non-stationary environments, where the underlying data distribution changes over time, is a central focus of data stream learning research. As reviewed in \cite{SUAREZCETRULO2023118934}, one of the prevailing strategies for handling concept drift is through the use of online learner ensembles. These methods leverage the diversity and modularity of multiple base learners, typically incremental models like the Hoeffding Tree \cite{domingos2000mining}, to create a system that is robust to shifts and drifts in the underlying data. The core objective of these ensembles is to resolve the stability-plasticity dilemma \cite{mermillod2013stability}: the model must be plastic enough to adapt to new concepts but stable enough to retain knowledge from past, still-relevant concepts.

Existing online ensemble methods can be broadly categorized into two groups:

\textbf{Active methods}: These approaches incorporate an explicit drift detection mechanism to signal a change in the data distribution. Upon detection, a corrective action is triggered. A state-of-the-art algorithm in this matter is the Adaptive Random Forest (ARF) \cite{Gomes2017AdaptiveClassification}, which keeps each of its base learners with a drift detector, such as the ADaptive WINdowing (ADWIN) algorithm \cite{bifet2007learning}. When a detector signals a "warning", it begins training a new "background" tree. If the detector confirms a "drift," the original tree is replaced by its background learner, allowing the ensemble to adapt quickly to abrupt drifts that impact its base learners. While highly effective, this approach can be computationally demanding, often requiring a large number of trees to maintain performance and diversity.

\textbf{Passive methods}: In contrast, passive methods adapt to drifts continuously without an explicit detection component. These algorithms often rely on dynamic weighting or instance-based adjustments. For example, OzaBag  \cite{oza2001online} uses online bagging with a drift detector to reset the worst-performing learner when a change is detected. Other methods, like Online Accuracy Updated Ensemble \cite{Brzezinski2014OAUE}, update learner weights based on their recent performance on blocks of data. Boosting-based approaches 
\cite{chen2012OSBoost} modify instance weights to focus on misclassified examples. Methods like Leveraging Bagging \cite{bifet2010leveraging} focus on increasing diversity by using a higher Poisson sampling rate, which has been shown to improve resilience to drift.

While these ensemble methods represent the state of the art, their resource consumption can be a significant drawback, particularly in resource-constrained environments like IoT gateways or edge devices \cite{ordonez2024adaptive}. Our work proposes an alternative to these large-scale ensembles by drawing inspiration from a different architectural paradigm.



\subsection{Mixture of Experts in Online Learning}

The Mixture of Experts (MoE) architecture, first introduced by Jacobs et al. \cite{jacobs1991adaptive}, is a model based on the divide-and-conquer principle \cite{liu2023diversifying}. It consists of two key components: a set of "expert" networks and a "gating network" (or router). The gating network learns to partition the input space, dynamically routing each input instance to the expert best suited to handle it. The final prediction is typically a weighted sum of the outputs from all experts, with weights determined by the gating network. The primary advantage of MoE is conditional computation; by activating only a sparse subset of experts for any given input, the model's capacity can be vastly increased without a proportional rise in computational cost.

\textbf{Traditional MoEs}: Traditionally, MoEs have been applied in batch learning settings to model complex functions or multi-modal data. More recently, they have seen a resurgence in large-scale deep learning use cases \cite{xue2024openmoe}, \cite{fedus2022switch}, \cite{dai2024deepseekMoE}. These approaches typically focus on scaling model capacity for static datasets, where the primary challenge is handling complex, high-dimensional data rather than temporal distribution shifts.
Their application to online or continual learning is less common but holds significant promise. In continual learning, MoEs have been explored as a way to mitigate catastrophic forgetting by allocating new tasks to new experts, thereby protecting previously acquired knowledge \cite{li2024theory}.

\textbf{MoEs in Continual/Streaming Learning}: MoE architectures have shown promise in continual learning by assigning new tasks to separate experts, helping prevent forgetting \cite{li2024theory}. However, these methods often assume task boundaries are known, which contrasts with streaming settings where concept drift is gradual and undefined. While recent work, such as Yang et al. \cite{10946581}, has applied MoEs to handle drift in specific domains, most approaches still rely on explicit drift detection or are domain-specific, limiting their adaptability to general streaming scenarios.



DriftMoE advances this direction by introducing a general-purpose, online co-training framework. Instead of static drift detectors or majority voting, a continuously trained neural router dynamically assigns data to incremental experts, enabling proactive, efficient adaptation in non-stationary environments.


\section{DriftMoE: Architecture and Training Procedure}

\subsection{Model Architecture}
The \textbf{DriftMoE} model implements a two-component mixture-of-experts (MoE) architecture \cite{shazeer2017outrageously}, composed of a set of streaming experts and a lightweight neural-network router. Let $\mathcal{E}=\{E_i\}_{i=1}^K$ denote the expert ensemble and $R_\theta$ the router parameterized by $\theta$. The router produces expert weights while each expert maintains its own incremental concept model. 

DriftMoE has two different variants, MoE-Data and MoE-Task, which differ only in the experts used, MoE-Data uses multi-class data experts updated in Top-K fashion and MoE-Task uses one-vs-rest binary experts updated every step. Both variants are trained online in a hybrid regime: experts update instance‐by‐instance, while the router updates in mini‐batches.

\paragraph{Experts.}
Each expert $E_i$ is instantiated as a Hoeffding tree \cite{domingos2000mining}, producing incremental updates on each observed instance. Two variants of expert configuration are supported:
\begin{enumerate}
\item \textbf{Data-mode, Multiclass Experts:} A set of $K$ Hoeffding trees, each issuing a $C$-way classification over the label set.

    \color{black}

  \item \textbf{Task‐mode experts:} One Hoeffding tree per class ($K=C$), each formulated as a binary classifier for its associated class. 
\end{enumerate}
Given an input $x_t$, expert $E_i$ outputs
\[
  \mathbf{p}_i(x_t) = \bigl[p_i^{(1)}, \dots, p_i^{(C)}\bigr]^\top,
  \quad
  p_i^{(c)} = P(y=c \mid x_t, E_i).
\]
In the task-mode variant, each expert reports a scalar $p_i(x_t)=P(y=i\mid x_t,E_i)$.

\paragraph{Router.}
The router $R_\theta$ is a three‐layer multilayer perceptron (MLP) that maps $x_t\in\mathbb{R}^d$ to expert logits
\[
  \mathbf{o}_t = R_\theta(x_t)\in\mathbb{R}^K.
\]
A softmax normalization yields gating weights
\[
  w_{t,i} = \frac{\exp(o_{t,i})}{\sum_{j=1}^K \exp(o_{t,j})},
  \quad
  \mathbf{w}_t = \mathrm{softmax}(\mathbf{o}_t).
\]
At inference time, the router may either (i) mix expert outputs via $\mathbf{w}_t$, or (ii) select the top‐$k$ experts by largest $w_{t,i}$.

\subsection{Joint Online Training}
\label{sec:training}

\paragraph{Expert Updates.}
\begin{itemize}
  \item \textbf{Multiclass mode (MoE-Data):}  
    For each incoming pair $(x_t,y_t)$, identify the top‐$k$ experts under $\mathbf{w}_t$, and update each selected $E_i$ using its Hoeffding tree rule:
    \[
      E_i.\mathrm{train}(x_t, y_t).
    \]
  \item \textbf{Task mode (MoE-Task):}  
    Define a binary label $y_{t,i} = \mathbf{1}[y_t = i]$ for each expert. Then update every $E_i$ on $(x_t,y_{t,i})$.
\end{itemize}

\paragraph{Router Updates.}
\begin{enumerate}
  \item \textbf{Correctness Mask.}  
    For each expert $E_i$, let
    \[
      \hat{y}_{t,i} = \arg\max_c p_i^{(c)}(x_t),
      \quad
      m_{t,i} = 
      \begin{cases}
        1, & \hat{y}_{t,i} = y_t,\\
        0, & \text{otherwise}.
      \end{cases}
    \]
    If $\sum_i m_{t,i}=0$, set $m_{t,y_t}=1$ to guarantee at least one positive target.
  \item \textbf{Mini‐Batch Optimization.}  
    Collect logits $\{\mathbf{o}_n\}_{n=1}^B$ and masks $\{\mathbf{m}_n\}_{n=1}^B$ into a batch of size $B$. Define the binary cross‐entropy loss
    \[
      \mathcal{L}_{\mathrm{BCE}}
      = -\frac{1}{B}\sum_{n=1}^B \sum_{i=1}^K \Bigl[
        m_{n,i}\,\log\sigma(o_{n,i})
        + (1-m_{n,i})\,\log\bigl(1-\sigma(o_{n,i})\bigr)
      \Bigr],
    \]
    where $\sigma(\cdot)$ denotes the sigmoid function applied to the router logits. Update $\theta$ via Adam with learning rate $\eta$:
    \[
      \theta \leftarrow \theta - \eta\,\nabla_\theta \mathcal{L}_{\mathrm{BCE}}.
    \]
\end{enumerate}

\subsection{Evaluation Metrics}
At each time step $t$, the system prediction $\hat{y}_t$ is obtained from the single expert $E_{i^*}$ with the highest router weight:
\[
  i^* = \arg\max_{i} w_{t,i},
  \quad
  \hat{y}_t = \arg\max_c p_{i^*}^{(c)}(x_t).
\]
Performance is assessed in a streaming (prequential) evaluation loop by comparing $\hat{y}_t$ with the true label $y_t$. We report accuracy, kappa m and kappa temporal metrics.
These metrics reflect the expert selected as the most reliable by the router in each instance.

\section{Experiments}

\subsection{Experimental Setup}
\label{sec:experiments}
We used six synthetic concepts, drift streams, and three real-world datasets that are 
popular in relevant data stream learning research \citep{Gomes2017AdaptiveClassification, bifet2007learning,bifet2010leveraging} offering a mix of abrupt and gradual drifts to showcase our model's performance in different scenarios 

LED streams \cite{Breiman1984LED} comprise 24 binary attributes of which 7 are informative and 17 are irrelevant. Drift is introduced at three points, at 250k instances, 500k instances, and 750k instances, respectively, drifting 3, 5, and 7 features. We use two variants: one with abrupt changes, where the transition width between regimes is 50 instances, and one with gradual changes, whose transition width is 50,000 instances.

SEA concepts are generated via the SEA generator \cite{street2001streaming}, which produces data streams with three continuous attributes, each ranging between 0 and 10. Only two determine class membership by following the following procedure: instances are uniformly sampled in the plane created by these and classified by comparing their sum against a block-specific threshold. A controlled 10 percent label noise may be injected, and class balance is enforced. We simulate three drifts between different thresholds to create an abrupt and gradual version.

RBF streams use the radial basis function generator, which places centroids at random positions, each associated with a standard deviation, weight, and class label. Instances are created by sampling a centroid (proportional to its weight) and offsetting it by a Gaussian perturbation. We used 50 centroids, and to model incremental drift, all centroids continuously move: in $RBF_m$ (moderate drift), the speed is set to 0.0001, while in $RBF_f$ (fast drift) it is set to 0.001.

Airlines show records of flight departure delays, and the objective is to predict whether a flight will be delayed or not, given seven mixed-type attributes and naturally occurring, season-dependent drift.

The Electricity dataset from the Australian New South Wales
Electricity Market offers 5-minute updates to wholesale price snapshots described by 8 attributes; market dynamics such as supply and demand introduce drift, and the goal is to predict whether the price will go up or down.

CoverType comprises remote-sensing measurements with 54 attributes and 7 forest classes, collected over the years and therefore affected by long-term distributional shifts.

All synthetic streams are produced with CapyMOA 1.3 \cite{capyMOA}, an extensible and efficient Python library that allows an integrated interface with external frameworks such as MOA \cite{MOA10} and Pytorch, and were generated by executing the exact command lines reported by Adaptive Random Forest (ARF) \cite{Gomes2017AdaptiveClassification}; this yields identical drift positions and allows for a direct comparison between performance metrics.

\begin{table}[hbt!]
\centering
\caption{Datasets.}
\label{tab:Datasets}
\begin{tabular}{lrrrrrrrr}
\toprule
Category      &    Stream & Instances &   Features & Classes \\
\midrule
Synthetic        &  LED (Abrupt) &  1,000,000 &  24 &    10 \\
        & LED (Gradual) &  1,000,000 &  24 &    10 \\
       & SEA (Abrupt) &  1,000,000 &  3 &    2  \\
      & SEA (Gradual) &  1,000,000 &  3 &    2  \\
      &  $RBF_m$ &  1,000,000 &  10 &    5 \\
       &  $RBF_f$ &  1,000,000 &  10 &    5 \\
Real          &  Airlines &  539,383 &  7 &    2 \\
         &  Electricty &  45,312  &  8 & 2 \\
          &  Cover Type &  581,012  &  54 & 7 \\
\bottomrule
\end{tabular}
\end{table}

These experiments have been run on a server with a 32-core AMD Ryzen Threadripper PRO 5975WX, 256 GB of RAM, and 2 x NVIDIA GeForce RTX 4090 GPUs.

Each individual expert in our MoE framework is implemented as a Hoeffding Tree with a fixed grace period of 50, meaning each leaf must observe 50 instances before evaluating potential splits. We use the variation with Naive Bayes at each leaf. We do not impose an explicit $m$ parameter to limit the number of splits; instead, split decisions are governed purely by the Hoeffding bound with our chosen confidence and grace period settings.

\begin{figure}[hbt]
  \centering
  \includegraphics[width=.85\linewidth]{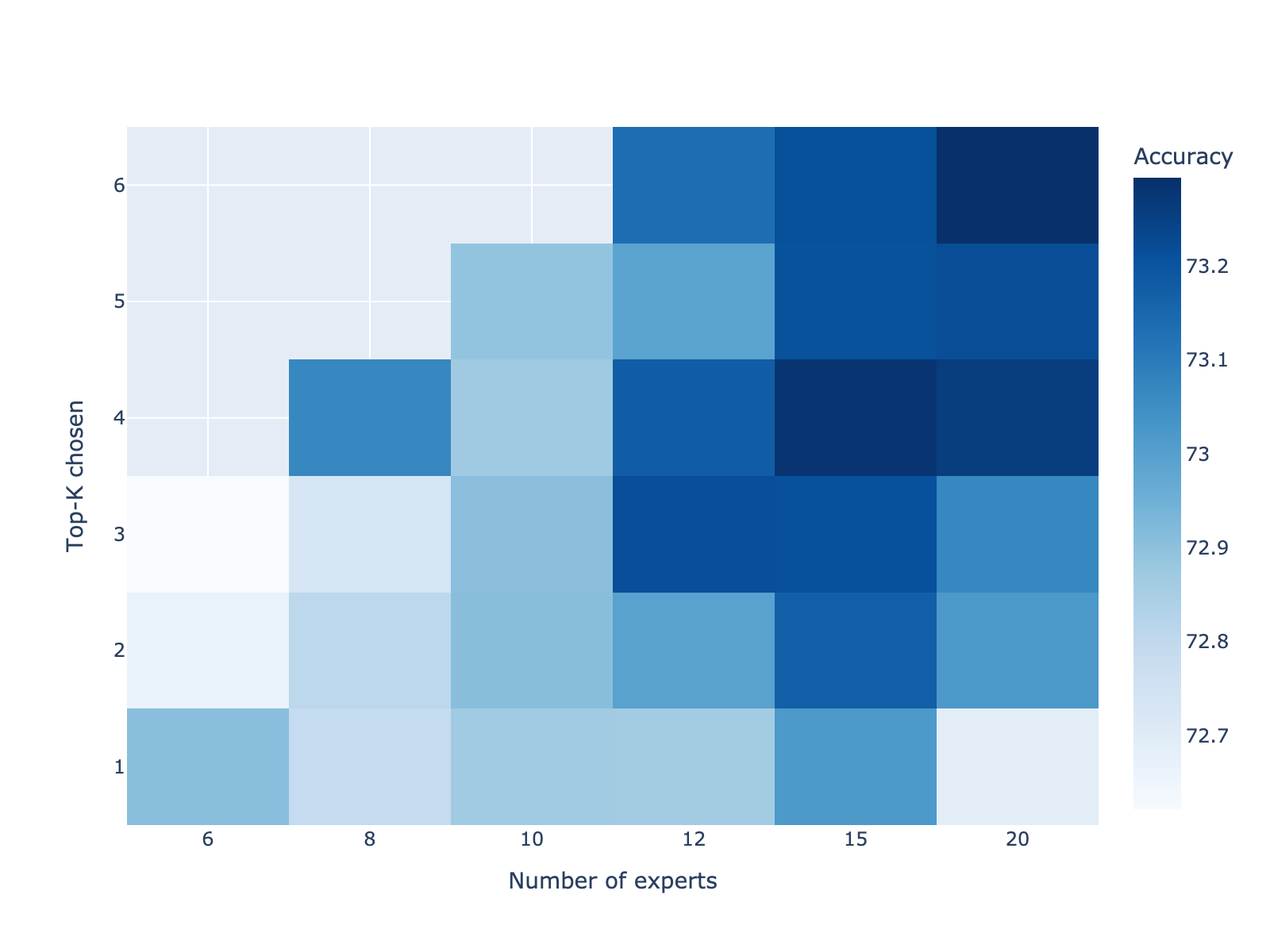}
  \caption{Grid search on the LED stream showing prequential accuracy
           as a function of the number of experts $N$ and the gating
           parameter $\text{Top-}K$.  The plateau at $(N{=}12,\text{Top-}K{=}3)$
           motivates our fixed configuration used in \ref{sec:experiments}}
  \label{fig:heatmap}
\end{figure}

Preliminary sweeps on the LED stream (Figure~\ref{fig:heatmap})
reveal a broad accuracy plateau for $12 \le K \le 20$ experts and
$3 \le k \le 5$ for the multi-class mode.  To keep runtime low and avoid per-dataset tuning we fix
$K = 12$ experts and $k = 3$ for all datasets.  This choice retains
at least $97\,\%$ of the peak accuracy while reducing compute
(by comparison with the 20-expert model) by roughly $30\,\%$.
All experts are initialised as identical, empty trees.

We employ the interleaved test-then-train procedure \cite{domingos2000mining}. And each stream is processed ten times with independent pseudo-random seeds; results are reported as mean ± standard deviation.

The same protocol is applied to both expert configurations: multi-class experts and task-specific (one-versus-rest) experts to guarantee a fair and consistent comparison across modes.

We adopt the single-model prequential protocol because it mirrors real-time deployment: the learner must predict each instance before observing its label and adapt continuously through drift, yielding a conservative, deployment-oriented performance estimate.

Beyond prequential accuracy, we measure Kappa-M, which corrects accuracy for chance agreement under class imbalance and drift \cite{bifet2015efficient}, and Kappa-Temporal, which discounts autocorrelation effects \cite{Zliobaite2015KappaT}

We compare our method against five widely-used adaptive stream ensembles, re-using the hyperparameters presented in their original studies, and they are assessed using the same prequential protocol as both DriftMoE variants.

\begin{itemize}
\item \textbf{Adaptive Random Forest (ARF)} \cite{Gomes2017AdaptiveClassification}: 100 Hoeffding-Tree base learners, each equipped with an \textsc{ADWIN} warning \& drift detector and a background tree; majority vote with tree-level weighting.
\item \textbf{OzaBag} \cite{oza2001online}: online bagging of 10 Hoeffding Trees using Poisson(1) re-sampling; when \textsc{ADWIN} signals drift, the worst tree is reset.
\item \textbf{OzaBoost} \cite{oza2001online} online boosting of 10 Hoeffding Trees using Poisson-weighted instance re-sampling.
\item \textbf{Online Smooth Boosting (SmoothBoost)} \cite{chen2012OSBoost}: stage-wise boosting of 30 Hoeffding Trees with smooth instance weighting that caps the influence of any single example and restarts learners when drift is detected.
\item \textbf{Leveraging Bagging (LevBag)} \cite{bifet2010leveraging}: bagging with amplified diversity via Poisson($\lambda=6$) re-sampling and output perturbation; the ARF benchmarks use an ensemble of 15 trees
\item \textbf{Streaming Random Patches} (SRP) \cite{gomes2019streaming} keeps an ensemble of incremental Hoeffding Trees, each grown on a Poisson-resampled stream and a fixed, globally random subset of features.
Every tree monitors its error with ADWIN; when drift is signalled the worst performer is discarded and restarted with a fresh feature patch.
\end{itemize}

These heavyweight baselines give the forthcoming results extra context: our router-based MoE matches or surpasses them while using far fewer trees. 




\subsection{Experimental Results}

\begin{figure}[!hbt]
  \centering
  \includegraphics[width=\textwidth]{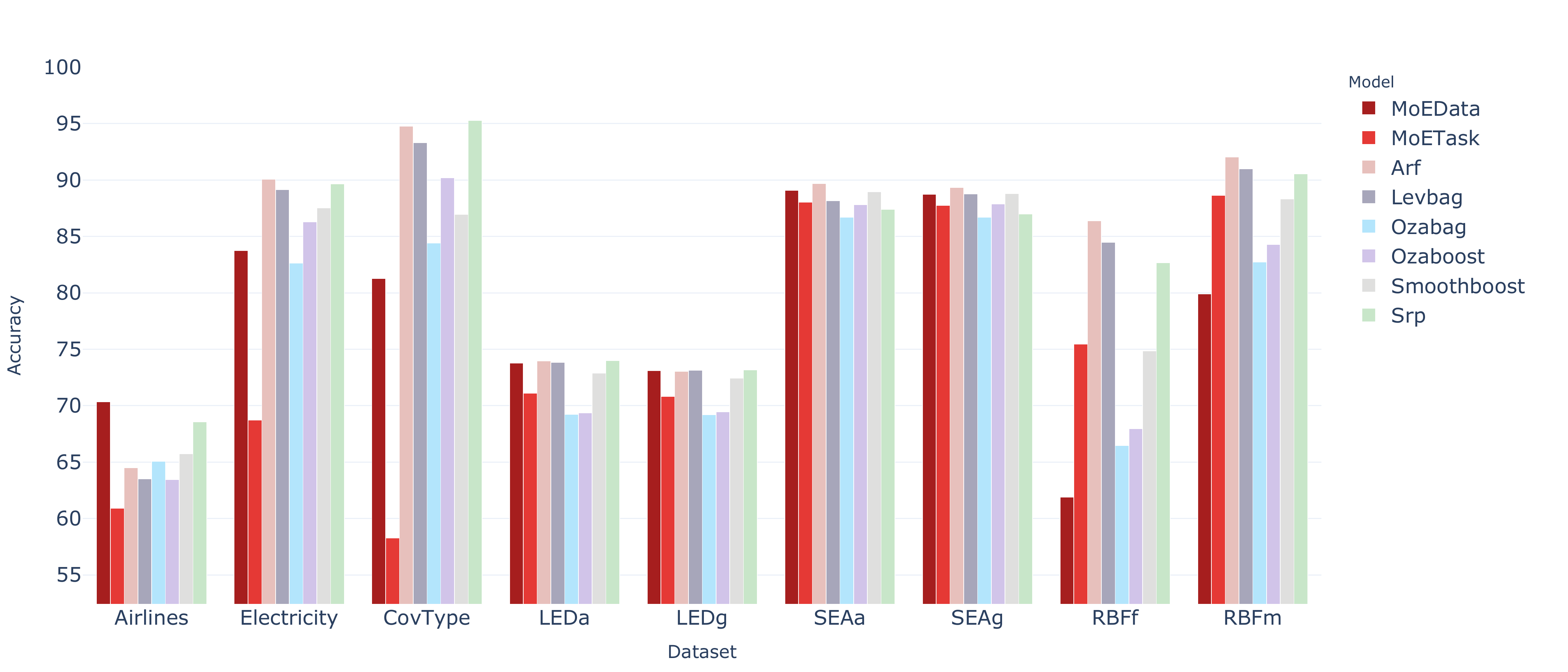}
  \caption{Prequential accuracy (\%) of baseline learners and DriftMoE variants across the nine benchmark datasets}
  \label{fig:accuracy}
\end{figure}

Table \ref{tab:accuracy_comparison} reports average accuracy over ten independent runs. \textbf{MoE-Task} ranks \emph{third} on RBF$_m$ and \emph{fourth} on RBF$_f$, confirming its aptitude for rapid high-frequency drift, whereas \textbf{MoE-Data} performs best on AIRL and remains consistenyl competitive on all LED and SEA variants. MoE-Data finishes last only on the on COVT yet still secures a podium position on AIRL, LED and SEA streams, illustrating the best overall stability trade-off.
In contrast, MoE-Task collapses on class-imbalanced streams such as ELEC and COVT, suggesting over-specialisation of its one-vs-rest experts. Figure \ref{fig:accuracy} displays these accuracy trends.

\begin{figure}[!hbt]
  \centering
  \includegraphics[width=\textwidth]{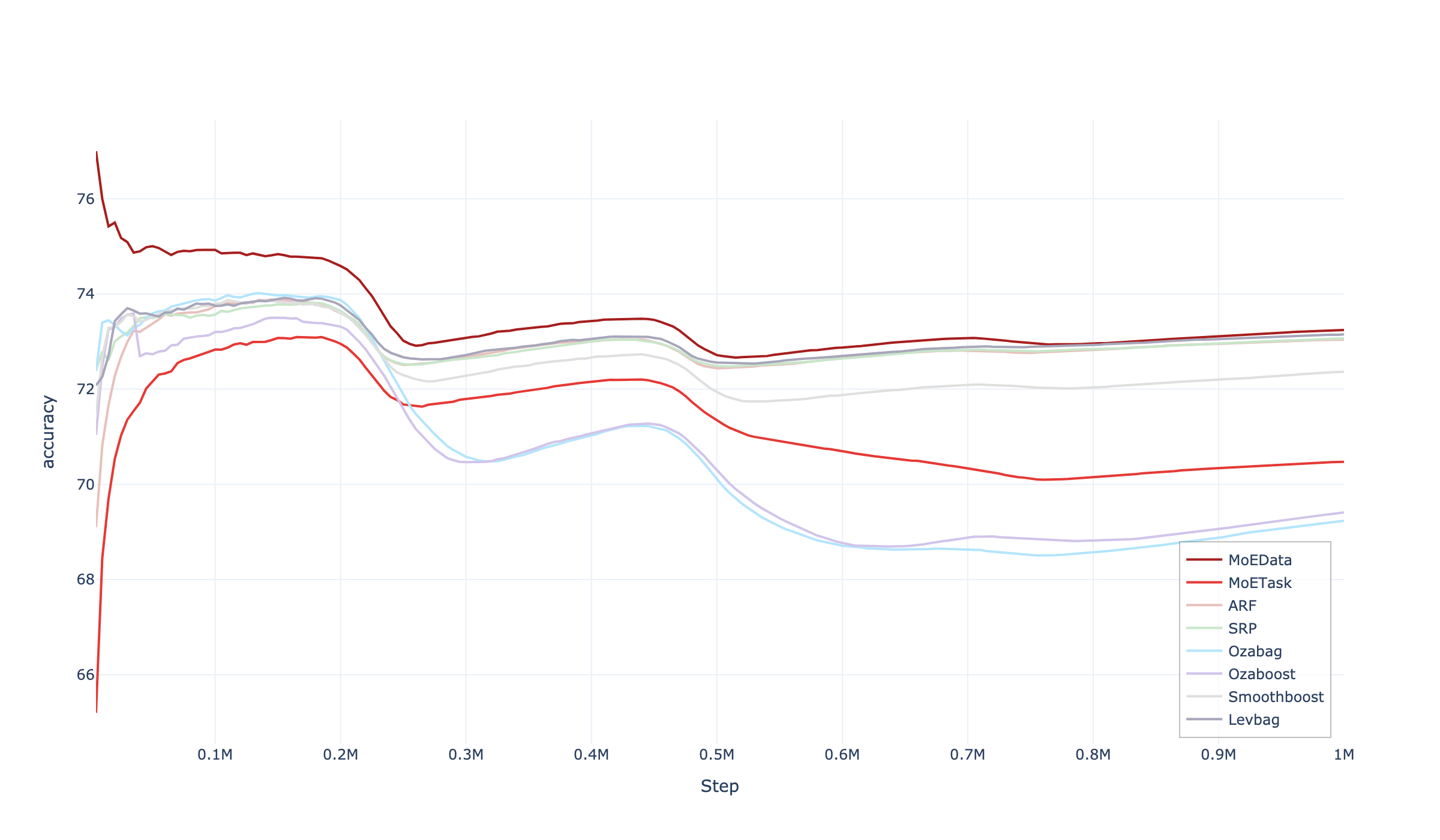}
  \caption{Accuracy over time plot for LED$_g$ dataset}
  \label{fig:accVStime}
\end{figure}

During experiment runs we observed that the router reacts to concept shifts with surprisingly little latency. Figure \ref{fig:accVStime} illustrates this for the LED$_g$ stream: after each scheduled drift, accuracy for the MoE‐Data rebounds at essentially the same recovery speed achieved by the much larger ADWIN-equipped ensembles (ARF and SRP), suggesting that the MoE design can match state-of-the-art drift-reaction speed while relying on an order-of-magnitude fewer base learners.

\begin{table}[ht]
\centering
\caption{Accuracy--baselines and DriftMoE variations}
\label{tab:accuracy_comparison}
\resizebox{\textwidth}{!} {
\begin{tabular}{lrrrrrrrr}
\toprule
Dataset & ARF & LevBag & OzaBag & OzaBoost & SmoothBoost & SRP & MoEData & MoETask \\
\midrule
Airlines   & 64.51 ± 0.02 & 63.50 ± 0.03 & 65.08 ± 0.02 & 63.45 ± 0.06 & 65.74 ± 0.00 & 68.55 ± 0.04 & \textbf{70.33 ± 0.18} & 60.92 ± 0.01 \\
CovType   & 94.78 ± 0.02 & 93.31 ± 0.02 & 84.41 ± 0.09 & 90.19 ± 0.23 & 86.95 ± 0.00 & \textbf{95.27 ± 0.01} & 81.28 ± 0.75 & 58.28 ± 0.31 \\
Electricity   & \textbf{90.08 ± 0.04} & 89.16 ± 0.07 & 82.63 ± 0.09 & 86.30 ± 0.40 & 87.51 ± 0.00 & 89.64 ± 0.12 & 83.76 ± 0.45 & 68.73 ± 0.85 \\
LED$_a$ & 73.96 ± 0.08 & 73.82 ± 0.08 & 69.22 ± 0.14 & 69.35 ± 0.34 & 72.88 ± 0.15 & \textbf{73.98 ± 0.08} & 73.77 ± 0.18 & 71.11 ± 0.54 \\
LED$_g$ & 73.04 ± 0.07 & 73.15 ± 0.08 & 69.21 ± 0.11 & 69.45 ± 0.19 & 72.44 ± 0.06 & \textbf{73.18 ± 0.07} & 73.11 ± 0.11 & 70.82 ± 0.38 \\
RBF$_f$ & \textbf{86.37 ± 1.30} & 84.47 ± 1.63 & 66.46 ± 1.89 & 67.95 ± 2.18 & 74.84 ± 1.48 & 82.66 ± 1.53 & 61.90 ± 0.20 & 75.45 ± 0.11 \\
RBF$_m$ & \textbf{92.04 ± 1.39} & 90.99 ± 1.37 & 82.73 ± 1.13 & 84.29 ± 1.72 & 88.31 ± 1.24 & 90.55 ± 1.23 & 79.89 ± 0.48 & 88.65 ± 0.07 \\
SEA$_a$ & \textbf{89.68 ± 0.04} & 88.17 ± 0.28 & 86.70 ± 0.06 & 87.81 ± 0.11 & 88.95 ± 0.07 & 87.39 ± 0.32 & 89.09 ± 0.05 & 88.04 ± 0.09 \\
SEA$_g$ & \textbf{89.33 ± 0.07} & 88.77 ± 0.08 & 86.69 ± 0.08 & 87.86 ± 0.09 & 88.79 ± 0.04 & 86.98 ± 0.64 & 88.74 ± 0.05 & 87.76 ± 0.04 \\
\bottomrule
\end{tabular}
}
\end{table}

Kappa M (Table \ref{tab:kappa_m}) and Kappa temporal (Table \ref{tab:kappa_temporal}) scores broadly reflect the accuracy findings. Across all streams the kappa metrics mirror the complementary strengths of the two variants while exposing their shared weakness on class-imbalanced data. A noteworthy detail is that, on all of the LED and SEA variants, MoE-Data exhibits the joint smallest gap between accuracy and Kappa M, implying that the multi-hot router is indeed capable of steering experts towards informative regions of the feature space.

\begin{table}[ht]
\centering
\footnotesize
\caption{Kappa M--baselines and DriftMoE variations}
\label{tab:kappa_m}
\resizebox{\textwidth}{!} {
\begin{tabular}{lrrrrrrrr}
\toprule
Dataset & ARF & LevBag & OzaBag & OzaBoost & SmoothBoost & SRP & MoEData & MoETask \\
\midrule
airl   & 20.33 ± 0.05 & 18.05 ± 0.06 & 21.60 ± 0.04 & 17.95 ± 0.14 & 23.09 ± 0.00 & 29.40 ± 0.10 & \textbf{33.39 ± 0.42} & 12.27 ± 0.02 \\
covt   & 89.82 ± 0.04 & 86.94 ± 0.05 & 69.58 ± 0.17 & 80.86 ± 0.44 & 74.54 ± 0.00 & \textbf{90.78 ± 0.03} & 63.46 ± 1.47 & 18.58 ± 0.60 \\
elec   & \textbf{76.63 ± 0.09} & 74.46 ± 0.17 & 59.07 ± 0.22 & 67.71 ± 0.94 & 70.57 ± 0.00 & 75.59 ± 0.28 & 61.74 ± 1.05 & 26.33 ± 2.00 \\
led\_a & 71.02 ± 0.09 & 70.86 ± 0.09 & 65.76 ± 0.16 & 65.89 ± 0.38 & 69.83 ± 0.17 & \textbf{71.04 ± 0.09} & 70.81 ± 0.20 & 67.84 ± 0.60 \\
led\_g & 70.00 ± 0.08 & 70.13 ± 0.09 & 65.73 ± 0.12 & 66.01 ± 0.22 & 69.33 ± 0.06 & \textbf{70.16 ± 0.08} & 70.08 ± 0.13 & 67.53 ± 0.42 \\
rbf\_f & \textbf{69.99 ± 3.38} & 64.18 ± 6.10 & 20.76 ± 12.78 & 20.83 ± 8.22 & 43.34 ± 5.66 & 61.30 ± 7.90 & 23.51 ± 0.40 & 50.71 ± 0.22 \\
rbf\_m & \textbf{81.43 ± 3.75} & 79.72 ± 3.34 & 55.80 ± 8.32 & 65.25 ± 4.47 & 72.42 ± 4.34 & 78.07 ± 4.27 & 59.64 ± 0.96 & 77.22 ± 0.14 \\
sea\_a & \textbf{74.17 ± 0.09} & 70.35 ± 0.67 & 66.68 ± 0.20 & 69.49 ± 0.30 & 72.32 ± 0.20 & 68.41 ± 0.77 & 72.79 ± 0.12 & 70.18 ± 0.22 \\
sea\_g & \textbf{73.26 ± 0.16} & 71.86 ± 0.19 & 66.65 ± 0.20 & 69.59 ± 0.22 & 71.89 ± 0.12 & 67.37 ± 1.60 & 71.90 ± 0.13 & 69.47 ± 0.10 \\
\bottomrule
\end{tabular}
}
\end{table}



\begin{table}[ht]
\centering
\caption{Kappa M--baselines and DriftMoE variations}
\label{tab:kappa_temporal}
\resizebox{\textwidth}{!} {
\begin{tabular}{lrrrrrrrr}
\toprule
Dataset & ARF & LevBag & OzaBag & OzaBoost & SmoothBoost & SRP & MoEData & MoETask \\
\midrule
airl   & 15.40 ± 0.06 & 12.98 ± 0.06 & 16.75 ± 0.04 & 12.88 ± 0.15 & 18.34 ± 0.00 & 25.03 ± 0.10 & \textbf{29.26 ± 0.44} & 6.84 ± 0.02 \\
covt   & -5.62 ± 0.41 & -35.44 ± 0.48 & -215.57 ± 1.74 & -98.56 ± 4.58 & -164.15 ± 0.00 & \textbf{4.31 ± 0.30} & -279.11 ± 15.28 & -744.69 ± 6.24 \\
elec   & \textbf{32.37 ± 0.27} & 26.09 ± 0.50 & -18.43 ± 0.63 & 6.58 ± 2.73 & 14.85 ± 0.00 & 29.37 ± 0.81 & -10.70 ± 3.04 & -113.17 ± 5.80 \\
led\_a & 71.06 ± 0.08 & 70.91 ± 0.09 & 65.80 ± 0.16 & 65.94 ± 0.37 & 69.87 ± 0.16 & \textbf{71.08 ± 0.09} & 70.85 ± 0.19 & 67.88 ± 0.60 \\
led\_g & 70.04 ± 0.08 & 70.17 ± 0.09 & 65.79 ± 0.12 & 66.05 ± 0.21 & 69.38 ± 0.06 & \textbf{70.20 ± 0.08} & 70.11 ± 0.13 & 67.56 ± 0.42 \\
rbf\_f & \textbf{72.29 ± 2.43} & 68.23 ± 3.72 & 30.54 ± 4.33 & 32.83 ± 2.12 & 48.88 ± 3.29 & 64.64 ± 4.01 & 23.87 ± 0.40 & 50.95 ± 0.22 \\
rbf\_m & \textbf{83.64 ± 2.81} & 81.68 ± 2.76 & 63.39 ± 3.66 & 68.23 ± 3.53 & 75.97 ± 2.71 & 80.60 ± 2.66 & 59.83 ± 0.95 & 77.33 ± 0.13 \\
sea\_a & \textbf{78.32 ± 0.08} & 75.12 ± 0.54 & 72.06 ± 0.15 & 74.39 ± 0.30 & 76.76 ± 0.17 & 73.46 ± 0.63 & 77.12 ± 0.10 & 74.92 ± 0.19 \\
sea\_g & \textbf{77.58 ± 0.14} & 76.42 ± 0.14 & 72.03 ± 0.16 & 74.49 ± 0.18 & 76.46 ± 0.10 & 72.62 ± 1.34 & 76.38 ± 0.11 & 74.33 ± 0.08 \\
\bottomrule
\end{tabular}
}
\end{table}

Finally, the findings can be summarised in two points: (1) joint router-expert training captures both sharp drift and regime shifts, and (2) both MoE variants remain sensitive to pronounced class imbalance, pointing the need for cost-sensitive losses or adaptive sampling in the future.

\section{Discussion}

Across all three metrics, accuracy, Kappa$_m$, and Kappa$_t$ DriftMoE is
competitive with state-of-the-art ensembles while using far fewer trees.
It tops the \texttt{Airlines} stream and stays within 2 pp of the leader on
both RBF streams, streams characterised by high-frequency drift. MoETask shows the highest reactivity, excelling in volatile conditions, while MoEData consistently performs near the top and emerges as the most robust across datasets. However, both variants show limitations under class imbalanced settings such as ELEC and COVT. These findings underscore three key takeaways: (1) adaptive synergy between router and experts enables efficient adaptation in dynamic environments, (2) sensitivity to imbalance remains a challenge for current router training strategies, and (3) variant selection matters where MoEData offers better stability and MoETask suits fast drift balanced scenarios. In summary, DriftMoE can match or outperform established ensembles under concept drift with far fewer resources, making it a promising alternative for scalable online learning, though future work should improve its handling of imbalance.

To our knowledge, this work represents one of the first implementations of Mixture of Experts applied directly data streams and online learning environments. We believe this is a promising direction with many unexplored design choices. A key area for improvement lies in enhancing the quality of the experts, particularly under more challenging and nonstationary data. We believe future work can benefit from more principled regime detection and dynamic expert allocation strategies, potentially improving accuracy and robustness across tasks. Exploring uncertainty-based routing, or drift-aware expert adaptation could further strengthen DriftMoE's applicability across domains.

\section{Conclusion}

This work introduced DriftMoE, a fully streaming Mixture of Experts framework designed to handle concept drift through online co-training of a neural router and a pool of incremental experts. We evaluated two variants: multi-class and task-specific, and showed that DriftMoE achieves competitive or superior performance to established ensemble baselines across synthetic and real-world streams, particularly in highly dynamic settings. The results demonstrate that DriftMoE maintains high adaptability with fewer learners, offering a more efficient alternative to large ensemble models like ARF. However, performance on class-imbalanced datasets highlights the need for better router calibration and expert assignment under skewed distributions. This opens several avenues for future work, including improving expert quality under harsh drift, enhancing the clustering of data regimes, and exploring more adaptive routing strategies. Overall, DriftMoE represents a flexible and efficient framework for streaming learning under nonstationarity, with strong potential for further improvement and extension.

\begin{credits}
\subsubsection{\ackname} 

This work was partially supported by the project ``Towards a
functional continuum operating system (ICOS)'' funded by the European Commission
under Project code/Grant Number 101070177 through the HORIZON EU program.

This work was partially supported by the project ``Open CloudEdgeIoT Platform Uptake in Large Scale Cross-Domain Pilots (O-CEI)'' funded by the European Commission
under Project code/Grant Number 101189589 through the HORIZON EU program.

Ricardo Simón Carbajo is the Director of Innovation and Development in Ireland's Centre for AI (CeADAR) and has been partially supported by the project ``Trustworthy Efficient AI for Cloud-Edge Computing (MANOLO)'' funded by the European Commission
under Project code/Grant Number 101135782 through the HORIZON EU program.

\subsubsection{\discintname}
The authors have no competing interests to declare that are
relevant to the content of this article. 
\end{credits}
%
%
%
%

\bibliographystyle{splncs04}

\bibliography{paper}





\end{document}